\documentclass[conference]{IEEEtran}
\IEEEoverridecommandlockouts
\usepackage{cite}
\usepackage{amsmath,amssymb,amsfonts}
\usepackage{algorithmic}
\usepackage{graphicx}
\usepackage{textcomp}
\usepackage{xcolor}
\usepackage{algorithm}
\usepackage{array}
\usepackage{textcomp}
\usepackage{stfloats}
\usepackage{url}
\usepackage{verbatim}
\usepackage{graphicx}
\usepackage{cite}
\usepackage{multirow}
\usepackage{enumitem}
\usepackage[breaklinks=true,colorlinks,linkcolor=blue,citecolor=blue,urlcolor=black]{hyperref}
\usepackage{flushend}
\usepackage{bbding}
\usepackage{xspace}
\usepackage{booktabs}
\usepackage{multirow}
\usepackage{amsmath,amssymb}
\usepackage{lineno}

\usepackage{times}           
\usepackage{makecell}

\usepackage{colortbl}
\definecolor{mygray}{rgb}{0.9,0.9,0.9}
\newcommand{\sexyname}{S$^3$KL\xspace}

\def\etal{{\em et al.}}

\DeclareMathAlphabet\mathbfcal{OMS}{cmsy}{b}{n}

\def\bmD{{\mathbfcal D}}

\def\0{{\bf 0}}
\def\1{{\bf 1}}

\def\bH{{\bf H}}
\def\bI{{\bf I}}

\def\eg{\emph{e.g.}}

\def\etal{\emph{et al.}}

\usepackage{amsmath}

\def\BibTeX{{\rm B\kern-.05em{\sc i\kern-.025em b}\kern-.08em
    T\kern-.1667em\lower.7ex\hbox{E}\kern-.125emX}}
\begin{document}

\title{Structure-aware Keypoint Localization for Videofluoroscopic Swallowing Study}


\author{
\IEEEauthorblockN{Kai Zhou$^{1*}$, Chuanshen Chen$^{1*}$, Runhao Zeng$^{2*}$, 
Meng Dai$^{3}$, Yifan Yang$^{4}$, Jinwu Hu$^{1}$, \\Daiyuan Li$^{1}$, Mingkui Tan$^{1\dagger}$, Fei Liu$^{1\dagger}$}

\IEEEauthorblockA{$^{1}$South China University of Technology, $^{2}$Shenzhen MSU-BIT University, \\$^{3}$The Third Affiliated Hospital of Sun Yat-sen University, $^{4}$Electric Power Research Institute, China South Grid}


\IEEEauthorblockA{\{kayjoe0723, chuanshenc888, youngyif1, fhujinwu, ldyabroad\}@gmail.com, zengrh@smbu.edu.cn}
\IEEEauthorblockA{daim@mail3.sysu.edu.cn, \{mingkuitan, feiliu\}@scut.edu.cn}

\thanks{$^*$Equal Contribution. $^\dagger$Corresponding Authors: Mingkui Tan, Fei Liu.}
\thanks{This work was partially supported by the National Natural Science Foundation of China (Nos. 62072190 and U24A20327), the Shenzhen Science and Technology Foundation (No. JCYJ20250604173210013), and the Key Scientific Research Project of the Department of Education of Guangdong Province (No. 2024ZDZX3012).}
}

\maketitle

\begin{abstract}
   Videofluoroscopic Swallowing Study (VFSS) is one of the gold standard for diagnosing swallowing disorders, providing dynamic X-ray imaging of the swallowing process. Automated kinematic analysis in VFSS relies fundamentally on precise anatomical keypoint localization. However, existing studies focus on limited keypoints (e.g., cervical vertebrae or the hyoid) and overlook critical regions such as the soft palate, while annotating only active swallowing segments and ignoring abundant non-swallowing data, resulting in poor data efficiency. Moreover, leveraging this unlabeled data via standard semi-supervised learning is suboptimal, as generic methods are prone to spatial bias. In medical X-rays with fixed layouts, models tend to memorize absolute coordinates rather than understanding anatomical structures. To tackle these challenges, we introduce VFSSKep, a novel dataset that extends annotations to the soft palate and incorporates large-scale unlabeled data. We further propose S$^3$KL, a Structure-aware Semi-Supervised Keypoint Localization framework designed to overcome spatial bias. It integrates a Structure-Aware Learning strategy to extract high-resolution structural cues for structure-aware representation learning, and a Structural Representation Consistency Learning strategy with block shuffling to enforce invariant structural recognition. Experiments show our method achieves state-of-the-art semi-supervised performance, even with unlabeled and 25\% labeled data surpassing fully supervised learning with 100\% labeled data. Code and data will be made publicly available at: \url{https://github.com/kaai520/S3KL}.
\end{abstract}


\section{Introduction}\label{sec:intro}

Swallowing disorders (dysphagia) \cite{rommel2016oropharyngeal} are a serious illness that frequently lead to life-threatening complications, such as severe malnutrition and aspiration pneumonia~\cite{carrion2019complications}.  Over 30\% of individuals aged 65 and older are affected by swallowing disorders~\cite{lee2019automatic}. The Videofluoroscopic Swallowing Study (VFSS) is one of the gold-standard diagnostic tools, providing dynamic X-ray imaging of the swallowing process~\cite{na2019thyroid}. Recently, clinical assessment has increasingly relied on quantitative kinematic analysis; for example, hyoid bone displacement is a key indicator for detecting aspiration and pharyngeal residues~\cite{kraaijenga2017hyoid}. This trend has driven the need for automated and precise localization of critical anatomical landmarks (e.g., the keypoints of the cervical vertebrae~\cite{zhang2021automatic} and hyoid bone~\cite{hsiao2023deep}).

\begin{figure}[t]
    \centering
    \includegraphics[width=0.48\textwidth]{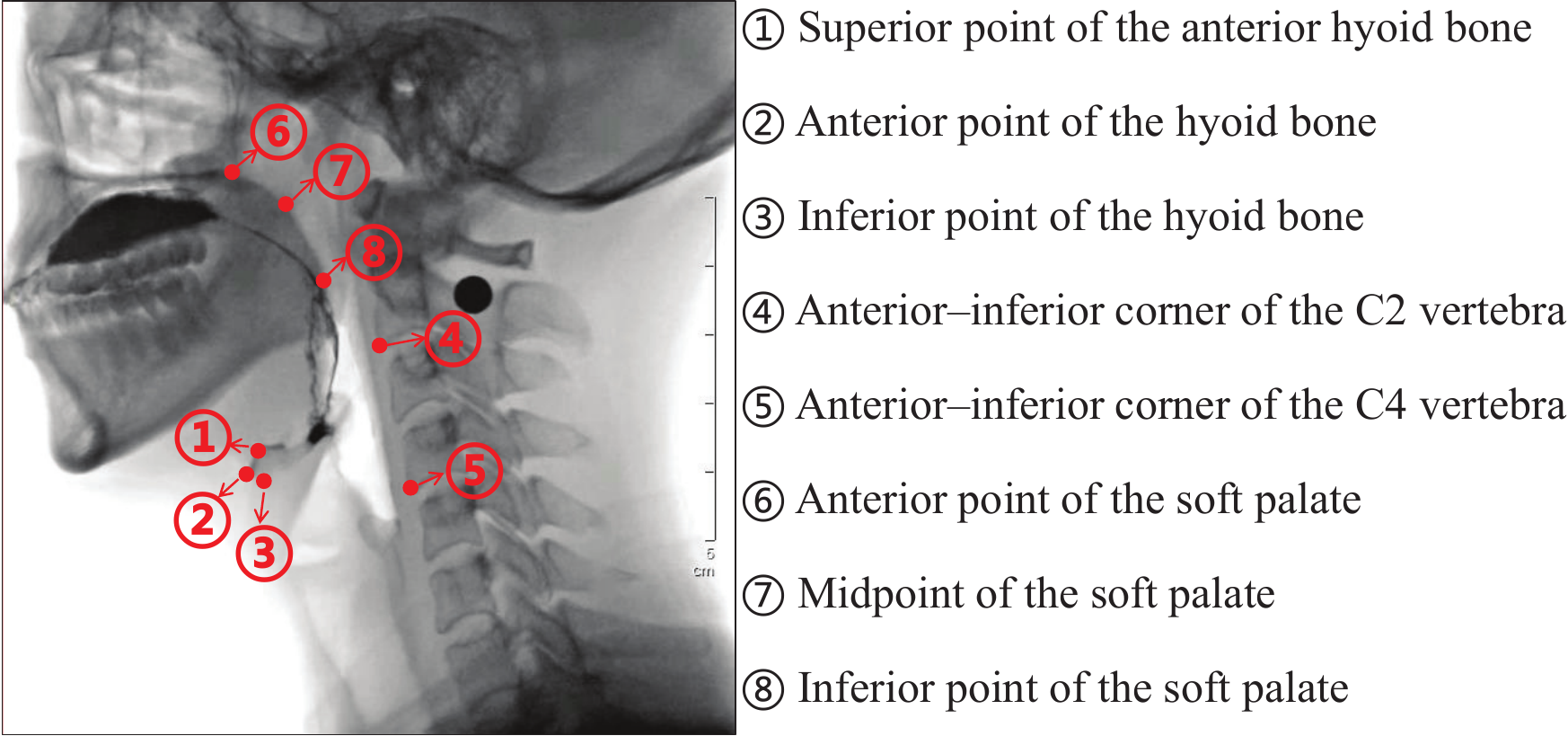}
    \vspace{-5pt}
    \caption{Illustration of the eight keypoints in our VFSSKep dataset.}
    \label{fig:8keypoints}
    \vspace{-10pt}
\end{figure}

Despite recent progress, existing VFSS keypoint (landmark) localization  benchmarks~\cite{zhang2021automatic,hsiao2023deep} are limited in anatomical scope and data utilization. \textit{First}, current datasets focus exclusively on rigid structures like the hyoid bone, overlooking the soft palate. This deformable stucture is critical for preventing nasal reflux~\cite{logemann2007swallowing}, yet its non-rigid motion makes manual annotation notoriously difficult~\cite{goldfield2010premature}, leaving it absent from automated systems. \textit{Second}, existing benchmarks suffer from poor data efficiency. While clinical videos are long (often minutes), annotations are typically restricted to brief ``active swallowing" windows (2-4 seconds), resulting in the exclusion of most non-swallowing frames and leaving a substantial amount of potentially informative anatomical data unused.

Leveraging this abundance of unlabeled data via Semi-Supervised Learning (SSL) is a natural solution, yet applying generic computer vision SSL paradigms to VFSS is non-trivial. Existing semi-supervised keypoint localization methods~\cite{xie2021empirical,Huang_2023_CVPR,ren2024g2lcps} typically rely on global geometric perturbations (\eg, rotation) to enforce consistency. In the context of medical X-ray imaging, which exhibits low tissue contrast and relatively fixed anatomical layouts, these methods are prone to shortcut learning. 
Previous models tend to memorize absolute keypoint coordinates (see Fig. \ref{fig:visplus}) rather than learning to distinguish the anatomical structure, leading to spatial bias that limits generalization diverse anatomical variations.

\begin{table}[t]
    \centering
    \caption{\textbf{Comparison of VFSS keypoint datasets.} Our VFSSKep is the first to include soft palate annotations and leverage large-scale unlabeled data for semi-supervised learning.}
    \label{tab:dataset}
    \setlength{\tabcolsep}{5pt} 
    \begin{tabular}{lcccc}
    \toprule
    Dataset & Year & \makecell{Anatomical\\Structure} & \makecell{Labeled\\Frames} & \makecell{Unlabeled\\Frames} \\
    \midrule
    Zhang \etal \cite{zhang2021automatic} & 2021 & Cervical Vertebrae & 60k & - \\
    Hsiao \etal \cite{hsiao2023deep} & 2023 & Hyoid + Vertebrae & 7k & - \\
    \midrule
    \rowcolor{gray!10} 
    \textbf{Ours (VFSSKep)} & \textbf{2025} & \makecell{\textbf{Hyoid + Vertebrae}\\ \textbf{+ Soft Palate}} & \textbf{14k} & \textbf{32k} \\
    \bottomrule
    \end{tabular}
\end{table}

\begin{figure}[t]
    \centering
    \includegraphics[width=0.41\textwidth]{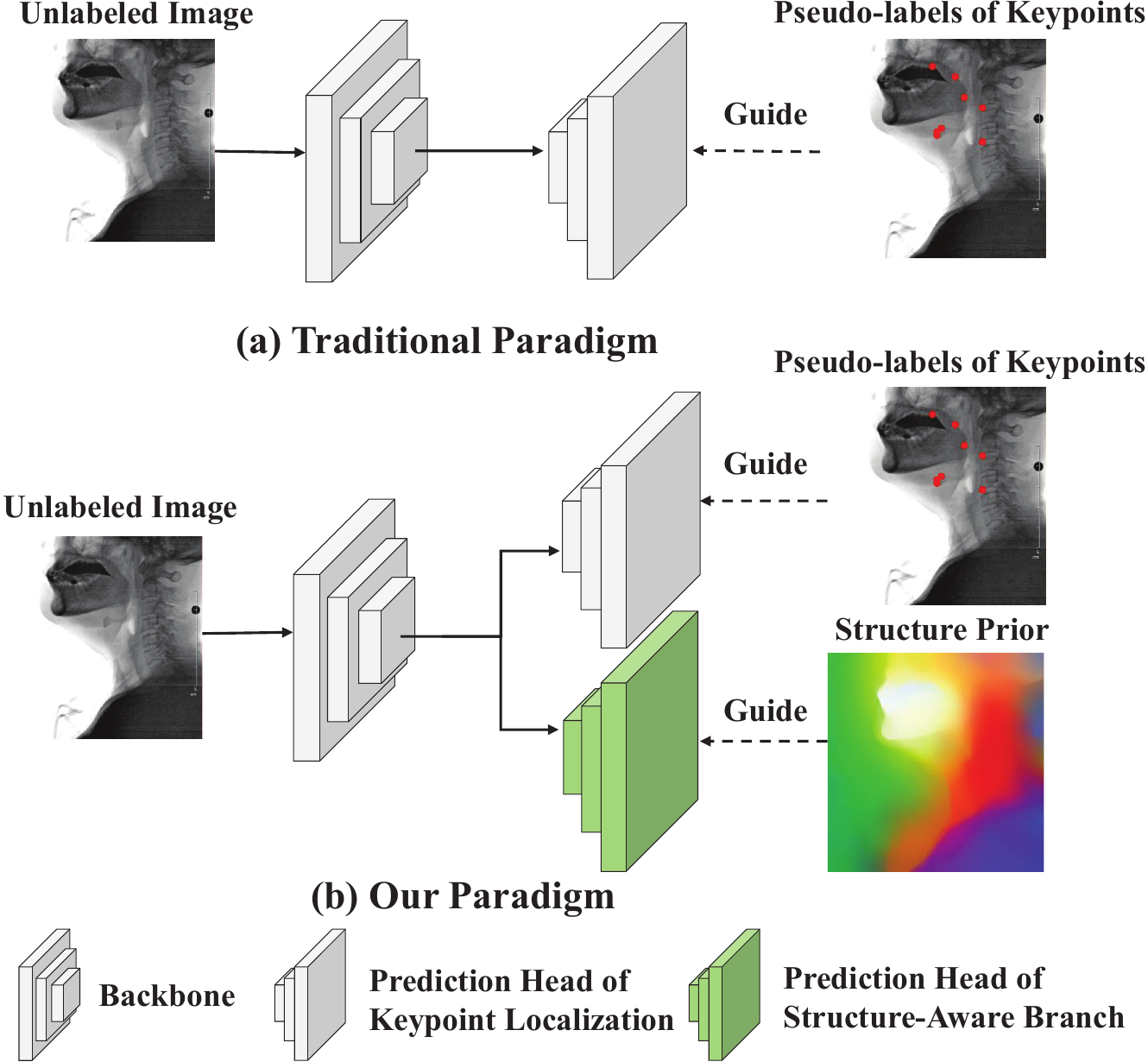}
    \caption{Comparison of the traditional (a) and our proposed (b) paradigms for utilizing unlabeled data in semi-supervised keypoint localization. Our paradigm leverages dense anatomical context to assist keypoint localization, learning structural priors without extra data or annotations.}
    \label{fig:ill}
    \vspace{-15pt}
\end{figure}

To bridge these gaps, we present a holistic solution comprising a new benchmark and a tailored learning framework. \textit{First}, we introduce VFSSKep, the first VFSS keypoint dataset to provide keypoint annotations for the deformable soft palate and incorporate large-scale non-swallowing segments as unlabeled data (see Fig.~\ref{fig:8keypoints} and Tab.~\ref{tab:dataset}). \textit{Second}, to effectively leverage these data, we propose a Structure-aware Semi-Supervised Keypoint Localization framework (\sexyname). Unlike conventional approaches, our method explicitly prevents shortcut learning by prioritizing anatomical understanding over spatial heuristics. We achieve this through two core paradigms: 1) A Structure-Aware Learning strategy. To compensate for the lack of dense medical annotations, we leverage the robust structure awareness of vision foundation models (\eg, DINOv2~\cite{oquab2024dinov2}). We design a module to recover high-resolution details from these general priors, guiding the encoder to perceive subtle anatomical structures (see Fig.~\ref{fig:ill}). 2) A Structural Representation Consistency Learning strategy. We introduce a ``Block Shuffle" operation to disrupt global spatial contexts at the input level. By restoring the feature positions and enforcing consistency with the original image, we compel the model to identify local anatomical parts independent of their absolute locations, significantly enhancing generalization capability.

Our contributions are as follows:
 
\begin{itemize}

\item We introduce VFSSKep, a novel VFSS keypoint dataset that extends anatomical scope to the soft palate keypoints and integrates large-scale unlabeled data, enabling data-efficient clinical analysis via semi-supervised learning.

\item We propose \sexyname framework, designed to overcome spatial bias through two core paradigms: a Structure-Aware Learning strategy that recovers high-resolution anatomical details from foundation model features to enforce dense structural alignment, and a Structural Representation Consistency Learning strategy for learning invariant features via input-level spatial disruption and feature-level restoration.

\item Experiments demonstrate the superiority of our approach. Notably, our \sexyname trained with only 25\% labeled data outperforms fully supervised baselines using 100\% labels, highlighting its exceptional data efficiency and generalization capability.



\end{itemize}

\section{VFSS Keypoint Dataset Construction}
\label{sec:dataset}

To address the limitations of existing benchmarks, we construct \textbf{VFSSKep}, a large-scale fluoroscopic video dataset tailored for semi-supervised keypoint localization.

\begin{figure}[tb]
    \centering
    \includegraphics[width=0.44\textwidth]{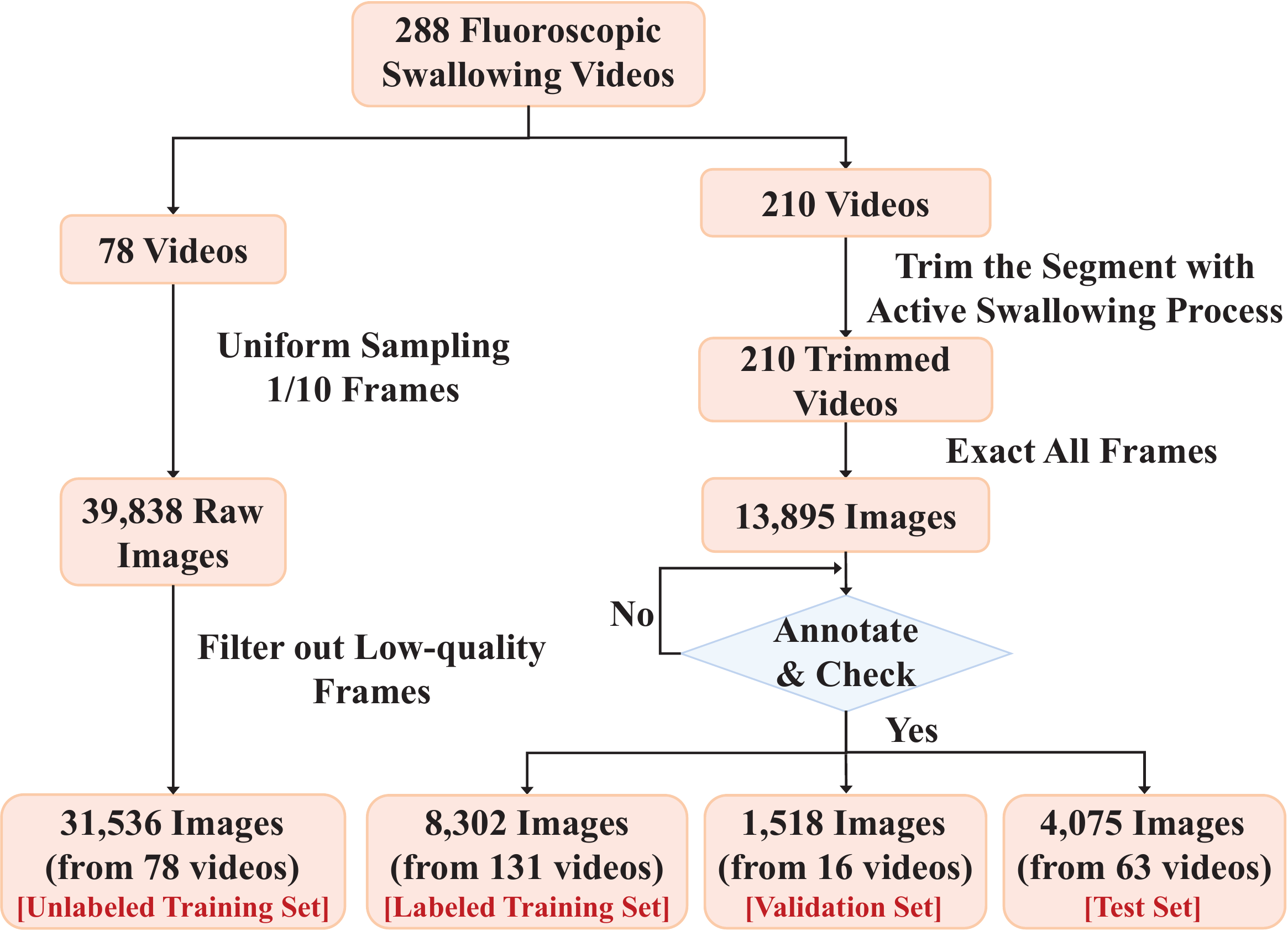}
    \caption{Summary of the construction of our VFSSKep dataset.}
    \label{fig:dataset}
    \vspace{-10pt}
\end{figure}

\textbf{Dataset Overview and Statistics.} The dataset was collected from a collaborating medical center using digital fluoroscopic systems. As shown in Fig.~\ref{fig:dataset}, it comprises 13,895 labeled frames extracted from active swallowing segments and 31,536 unlabeled frames from non-swallowing segments.
For the labeled data, we follow a subject-independent split ratio of roughly 6:1:3, resulting in 8,302 training, 1,518 validation, and 4,075 test images.
Crucially, distinct individuals are strictly separated across subsets to prevent data leakage.
The unlabeled set, derived from the remaining videos, captures diverse static and motion patterns to support semi-supervised learning.
\textit{Detailed acquisition protocols, filtering strategies, and demographic statistics are provided in the \textbf{Appendix}.}

\textbf{Keypoint Definition.} As shown in Fig.~\ref{fig:8keypoints}, we annotate \textit{eight anatomical landmarks} essential for swallowing kinematics: the superior, anterior, and inferior points of the hyoid bone; the anterior–inferior corners of C2 and C4 vertebrae; and three points on the soft palate (anterior, midpoint, inferior). Annotations were performed by clinical experts with strict multi-step verification.

\textbf{Comparison with Existing Datasets.} VFSSKep differs in two aspects:
\textit{1) Deformable structures:} Unlike prior datasets~\cite{zhang2021automatic,hsiao2023deep}, it includes soft palate keypoints, enabling oral-phase analysis.
\textit{2) Semi-supervised design:} It provides abundant unlabeled non-swallowing frames, reflecting clinical label scarcity and offering a semi-supervised setting.

\begin{figure*}[tb]
    \centering
    \includegraphics[width=1.0\textwidth]{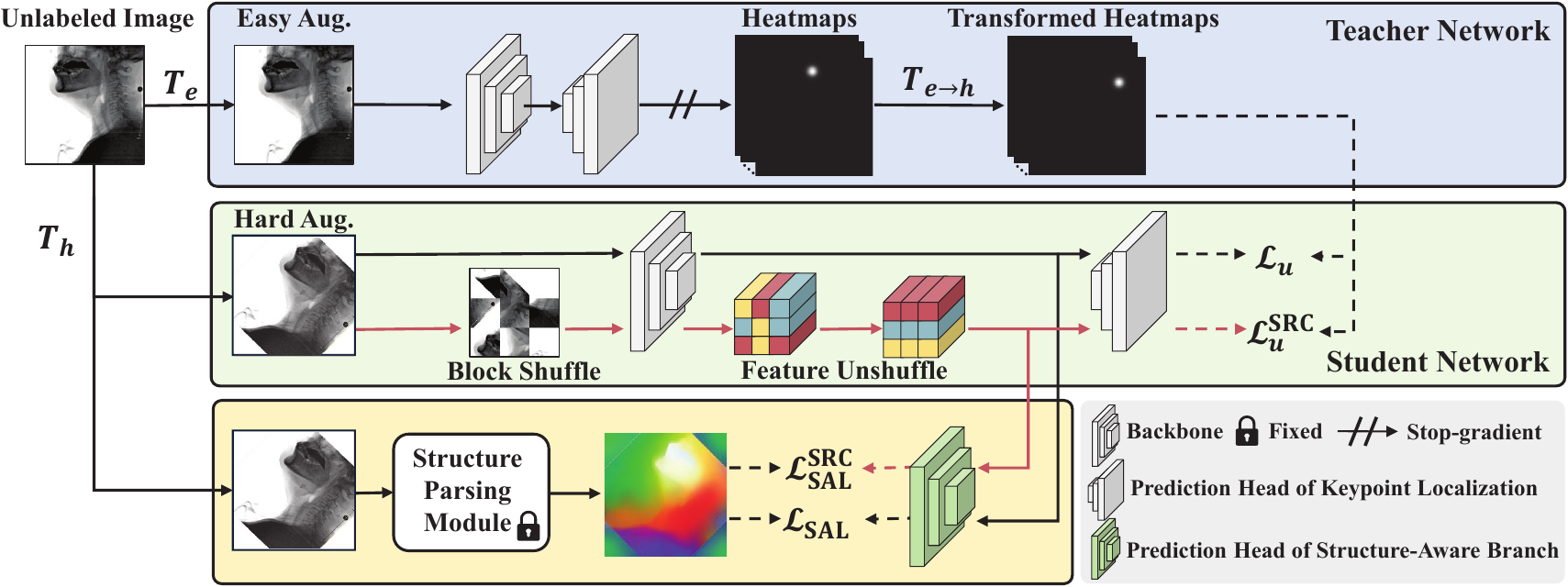}
    \caption{An overview of our proposed \sexyname framework. Our proposed \sexyname framework applies easy augmentation $T_e$ to an unlabeled image for teacher predictions, transformed via easy-to-hard mapping $T_{e \to h}$ as pseudo-labels for the student. Beyond keypoint supervision, we propose a Structure-Aware Learning (SAL) strategy to leverage dense anatomical cues, assisting keypoint localization by aligning the auxiliary prediction head output with structural prior from the Structure Pairing Module. The output from the Structure Parsing Module is visualized using PCA~\protect\cite{pearson1901liii}. In addition, we introduce a Structural Representation Consistency (SRC) Learning strategy, where the Block Shuffle operation disrupts spatial structure, encouraging the model to maintain consistent representations despite such perturbations. The training process of labeled images is omitted for simplicity.}
    \label{fig:framework}
    \vspace{-10pt}
\end{figure*}

\section{Method}\label{sec:method}

\subsection{Preliminaries and Method Overview}

\textbf{Problem Formulation.} 
We address the problem of semi-supervised keypoint localization. Given a labeled dataset $\bmD_s=\{(\bI_i^s, \bH_i^s)\}_{i=1}^{N_s}$ and a larger unlabeled dataset $\bmD_u=\{\bI_i^u\}_{i=1}^{N_u}$, our goal is to train a robust model $f_\theta$ by leveraging both sets. For each image $\bI$, the ground truth $\bH$ consists of $K$ Gaussian heatmaps, each indicating the probability distribution of a specific anatomical keypoint.

\textbf{Standard SSL Paradigm.} 
Existing methods~\cite{xie2021empirical,ren2024g2lcps} typically adopt a Teacher-Student framework. The student network is trained on labeled data via a supervised loss $\mathcal{L}_{s}$ and on unlabeled data via an unsupervised loss $\mathcal{L}_{u}$. Taking SemiPose~\cite{xie2021empirical} as a baseline, the learning objective is:
\begin{align}\label{eq:baseline}
    \mathcal{L}_{base} &= \mathcal{L}_{s} + \lambda \mathcal{L}_{u}, \\
    \mathcal{L}_{s} &= \mathbb{E}_{(\mathbf{I}^s, \mathbf{H}^s) \sim \bmD_s} [\| f_{\theta}(T_e(\mathbf{I}^s)) - T_e(\mathbf{H}^s) \|^2 ], \\
    \mathcal{L}_{u} &=\mathbb{E}_{\mathbf{I}^u \sim \bmD_u}[ \| f_{\theta}(T_h(\mathbf{I}^u)) - T_{e\to h}(f_{\theta'}(T_e(\mathbf{I}^u))) \|^2 ],
\end{align}
where $T_e$ denotes an easy augmentation. Consistency is enforced between the student under a hard augmentation $T_h$ and the teacher under $T_e$, with $T_{e\to h}$ aligning the teacher output to the student view. Here, $\theta$ and $\theta'$ denote the student and teacher parameters, and $\lambda_1$, $\lambda_2$ balance the loss terms.

\textbf{Method Overview.}
Due to the fixed anatomical layouts and potential spatial bias in VFSS X-ray imaging, learning robust features from unlabeled data without shortcut learning is a highly challenging task. To address this, we propose a structure-aware semi-supervised keypoint localization framework, namely \sexyname. As illustrated in Fig.~\ref{fig:framework}, the framework comprises two core parts: 
1) \textbf{Structure-Aware Learning}. It is implemented through an auxiliary branch that extracts dense structural representations from the encoder's features. This branch is supervised by a reference signal generated through our proposed Structure Parsing Module, which produces meaningful anatomical patterns from the input image without requiring additional annotations. 
2) \textbf{Structural Representation Consistency Learning}. We design a ``Block Shuffle'' mechanism to explicitly disrupt global spatial contexts at the input level. By compelling the model to restore feature consistency under such perturbations, this module enforces the learning of invariant anatomical representations, significantly enhancing generalization capability.

\subsection{Structure-Aware Learning} \label{sec:SAB}

\begin{figure}[tb]
    \centering
    \includegraphics[width=0.48\textwidth]{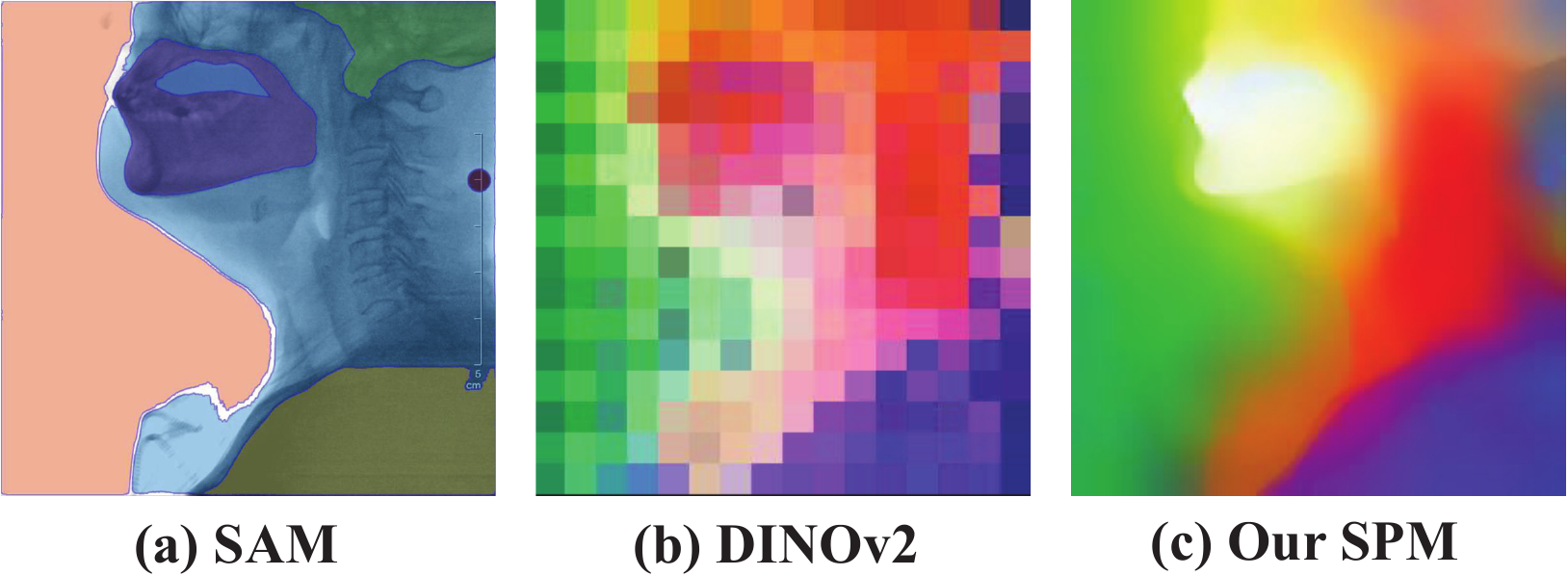}
    \caption{(a)  X-ray image segmented by SAM~\cite{kirillov2023segment} with regions highlighted in different colors, but without semantic labels. (b) Output features from DINOv2~\cite{oquab2024dinov2} for the same image, visualized by mapping PCA results to RGB channels. (c) Output features from our Structure Parsing Module (SPM), also visualized using PCA mapping.}
    \label{fig:vis}
    \vspace{-10pt}
\end{figure}

To prioritize anatomical understanding, it is essential to enforce dense structural constraints on the model. While prior studies~\cite{xia2017joint} typically achieve this by introducing auxiliary semantic segmentation tasks, obtaining such pixel-wise annotations for VFSS is prohibitively costly. Defining anatomically meaningful regions is also ambiguous due to the complexity of soft tissues. Although auto-segmentation methods such as SAM~\cite{kirillov2023segment} and MedSAM~\cite{ma2024segment} produce coarse contours, they lack VFSS-specific semantics and often fail to distinguish functionally relevant regions (Fig.~\ref{fig:vis}(a)).

\textbf{Structure Parsing Module.} To address this, we propose a Structure Parsing Module (SPM) that identifies critical anatomical regions without manual annotations. We leverage large-scale self-supervised models (e.g., DINOv2~\cite{oquab2024dinov2}) for contextual priors, but their limited output resolution causes blurred boundaries and loss of detail (Fig.~\ref{fig:vis}(b)). We therefore incorporate FeatUp~\cite{fu2024featup} to recover high-resolution, semantically consistent features. As shown in Fig.~\ref{fig:vis}(c), the enhanced SPM yields clear anatomical structures such as the soft palate (white), hyoid bone (yellow), and vertebrae (red) that align with the target keypoints, improving interpretability and providing strong structural guidance without medical annotations.

\textbf{Structure-Aware Branch.} 
To effectively distill these dense priors into the localization model, we propose a Structure-Aware Branch (SAB). Specifically, we append an auxiliary prediction head $f_{\theta,s}$ to the shared backbone $f_{\theta,\text{enc}}$. Unlike traditional multi-task learning that relies on discrete semantic segmentation labels~\cite{xia2017joint}, our SPM yields continuous, high-resolution structural features. Accordingly, we replace the conventional cross-entropy loss with a cosine similarity loss to align the student representations with the structural targets provided by SPM. The loss function is defined as:
\begin{equation}\label{eq:cos}
    \mathcal{L}_{\text{cos}}(\mathbf{x}) = \mathbb{E}_{(i,j)} \left[ \mathbb{I}(\text{sim}_{ij} \le \alpha) (1 - \text{sim}_{ij}) \right],
\end{equation}
where $\text{sim}_{ij} = \cos(f_{\theta,s}(\mathbf{x})_{ij}, f_{\text{SPM}}(\mathbf{x})_{ij})$ denotes the cosine similarity between the SAB prediction and SPM target at location $(i,j)$. The indicator function $\mathbb{I}(\cdot)$ works with the threshold $\alpha$ to focus learning on misaligned regions.

Finally, we apply this constraint to unlabeled data under hard augmentation to enforce robust structural representation learning. The overall Structure-Aware Learning loss is:
\begin{equation}\label{eq:SAL}
    \mathcal{L}_{\text{SAL}} = \mathbb{E}_{\mathbf{I}^u \sim \bmD_u} [\mathcal{L}_{\text{cos}}(T_h(\mathbf{I}^u))].
\end{equation}

\subsection{Structural Representation Consistency Learning} \label{sec:SRC}

To explicitly disrupt the spatial bias where models memorize absolute coordinates, we introduce a Structural Representation Consistency (SRC) Learning strategy. Unlike photometric augmentations that only alter pixel values, our approach destroys global geometric context via a Block Shuffle (BS) operation, compelling the encoder to recognize anatomical structures based on local visual patterns rather than absolute positions.

Formally, given an input image $\mathbf{x}$, we partition it into a grid of non-overlapping blocks and apply a random permutation $T_{\text{BS}}$, yielding a shuffled image $\mathbf{x}_s = T_{\text{BS}}(\mathbf{x})$. This shuffled input is processed by the encoder $f_{\theta,\text{enc}}$. Crucially, unlike pretext tasks that predict permutation indices~\cite{noroozi2016unsupervised}, our goal is feature invariance. Therefore, we apply an \textit{inverse} operation $T_{\text{US}}$ (Unshuffle) to the encoded features, restoring the original spatial order before the prediction heads:
\begin{equation}
    \mathcal{F}_o = T_{\text{US}}(f_{\theta,\text{enc}}(T_{\text{BS}}(\mathbf{x}))).
\end{equation}

This process forces the encoder to generate logically consistent features even when the global spatial layout is corrupted, effectively enforcing \textit{local-to-global} anatomical reasoning.

We integrate this mechanism into the semi-supervised consistency framework. The student model processes the block-shuffled image, and its spatially restored features are forced to align with the teacher's stable predictions. The modified unsupervised loss $\mathcal{L}_{u}^{\text{SRC}}$ is formulated as:
\begin{equation}\label{eq:SRC}
    \mathcal{L}_{u}^{\text{SRC}} =\mathbb{E}_{\mathbf{I}^u \sim \bmD_u}[ \| g_{\theta}(T_h(\mathbf{I}^u)) - T_{e\to h}(f_{\theta'}(T_e(\mathbf{I}^u))) \|^2 ],
\end{equation}
where $g_{\theta}(\cdot) = f_{\theta,k} \circ T_{\text{US}} \circ f_{\theta,\text{enc}} \circ T_{\text{BS}}$ represents the student's forward pass with Block Shuffle. Here, $f_{\theta,k}$ denotes the keypoint prediction head, and $T_{\text{US}}$ restores the spatial order of features extracted by the encoder $f_{\theta,\text{enc}}$. Note that this strategy is similarly applied to the Structure-Aware Learning loss ($\mathcal{L}_{\text{SAL}}$) to further robustify structural feature extraction.

\subsection{Overall Learning Objectives}

The total learning objective combines the supervised loss, standard unsupervised loss, and our proposed structure-aware regularization terms:
\begin{equation}
   \mathcal{L}=\lambda_1\mathcal{L}_{s}+\lambda_2(\mathcal{L}_u+\mathcal{L}_{u}^{\text{SRC}})+\lambda_3(\mathcal{L}_{\text{SAL}}+\mathcal{L}_{\text{SAL}}^{\text{SRC}}),
\end{equation}
where $\lambda_{1,2,3}$ are balancing coefficients. Here, $\mathcal{L}_{u}^{\text{SRC}}$ and $\mathcal{L}_{\text{SAL}}^{\text{SRC}}$ denote the consistency and structure-aware losses applied to the block-shuffled branch, respectively.

\textbf{Dual Networks Learning.} 
Following SemiPose~\cite{xie2021empirical}, we employ two networks $f_{\theta}$ and $f_{\xi}$ to mutually supervise each other. Instead of a standard EMA teacher, each network acts as a pseudo-label generator for the other on unlabeled data. Specifically, for the consistency terms ($\mathcal{L}_u$ and $\mathcal{L}_{u}^{\text{SRC}}$), we compute the loss by aligning $f_{\theta}$'s predictions with the fixed pseudo-labels from $f_{\xi}$ (and vice versa). This interaction prevents coupling errors and stabilizes training. Final predictions are obtained by fusing the outputs of both networks.

\section{Experiments and Results}

\subsection{Experimental Settings}
\label{sec:exp_setting}
\textbf{Datasets and Metrics.} We conduct experiments on our VFSSKep Dataset (Sec.~\ref{sec:dataset}), as existing benchmarks~\cite{zhang2021automatic,hsiao2023deep} are unavailable. We use 25\%, 50\%, and 100\% of the labeled training data, each combined with the full unlabeled set, and evaluate on the VFSSKep test set. Evaluation relies on two metrics: Probability of Correct Keypoint (PCK) and Mean Euclidean Distance (MED). For PCK, we adopt a strict threshold (PCK@0.02), considering a keypoint correct if it falls within 2\% of the image diagonal. MED measures the average physical distance (mm) for complementary precision analysis.

\textbf{Implementation Details.} We employ SimpleBaseline~\cite{xiao2018simple} with a ResNet18~\cite{he2016deep} backbone ($256 \times 256$) as the backbone. The Structure Parsing Module integrates a fixed DINOv2 with JBU upsamplers~\cite{fu2024featup}. Models are trained for 100 epochs (batch size 64) using Adam with an initial learning rate of $10^{-3}$, decayed at epochs 60 and 90. Following SemiPose~\cite{xie2021empirical}, we apply easy (rotation $\pm30^\circ$) and hard (rotation $\pm60^\circ$ with our Block Shuffle) augmentations to the teacher and student, respectively. Loss weights are set to $\lambda_1=0.5$, $\lambda_2=1.0$, and feature alignment $\lambda_3=10^{-3}$, with a similarity threshold $\alpha=0.85$. \textit{Architectural details of SAB and SPM, full hyperparameter analysis, and extended experiments (e.g., backbone variants) are provided in the \textbf{Appendix}.}


\textbf{Baselines.} We compare \sexyname against methods representing key milestones in the field: the supervised lower bound SimpleBaseline~\cite{xiao2018simple}, the classic semi-supervised method SemiPose~\cite{xie2021empirical}, and the recent state-of-the-art G2LCPS~\cite{ren2024g2lcps}. \textit{Comprehensive related works are discussed in the \textbf{Appendix}.}

\begin{table*}[tb]
\renewcommand{\arraystretch}{1.0}
\renewcommand{\tabcolsep}{11pt}
\begin{center}
\caption{Comparison with methods using different proportions of labeled data. PCK@0.02 (\%) and MED (mm) are reported. Our \sexyname is plug-and-play, where variations in unsupervised loss $\mathcal{L}_u$ across different methods do not affect its integration. Best results are in \textbf{bold}, and \textcolor{blue}{blue numbers} show \textbf{relative} improvements over SemiPose and G2LCPS.}
\label{tab:method_comparison}
\begin{tabular}{lllllll}
\toprule
\multirow{2}{*}{Method} & \multicolumn{2}{c}{25\%} & \multicolumn{2}{c}{50\%} & \multicolumn{2}{c}{100\%} \\
\cmidrule(lr){2-3} \cmidrule(lr){4-5} \cmidrule(lr){6-7}
                        & PCK@0.02 (\%)↑ & MED (mm)↓ & PCK@0.02 (\%)↑ & MED (mm)↓ & PCK@0.02 (\%)↑ & MED (mm)↓ \\
\midrule
Supervised~\cite{xiao2018simple}          & 63.2 & 4.34  & 73.5 & 3.47  & 75.6 & 3.29 \\
\midrule
SemiPose~\cite{xie2021empirical}          & 77.2 & 3.63  & 79.0 & 3.14  & 81.9 & 2.65 \\
+\sexyname (Ours)                          & 78.8 (\textcolor{blue}{↑2.1\%}) & 2.88 (\textcolor{blue}{↓20.7\%}) 
                                          & 82.1 (\textcolor{blue}{↑3.9\%}) & 2.64 (\textcolor{blue}{↓15.9\%})
                                          & 83.6 (\textcolor{blue}{↑2.1\%}) & 2.48 (\textcolor{blue}{↓6.4\%}) \\
\midrule
G2LCPS~\cite{ren2024g2lcps}               & 78.2 & 2.98  & 81.0 & 2.76  & 83.1 & 2.54 \\
+\sexyname (Ours)                 & \textbf{79.4} (\textcolor{blue}{↑1.5\%}) & \textbf{2.87} (\textcolor{blue}{↓3.7\%})
                                          & \textbf{82.9} (\textcolor{blue}{↑2.3\%}) & \textbf{2.57} (\textcolor{blue}{↓6.9\%})
                                          & \textbf{83.8} (\textcolor{blue}{↑0.8\%}) & \textbf{2.46} (\textcolor{blue}{↓3.1\%}) \\
\bottomrule
\end{tabular}

\end{center}
\vspace{-15pt}
\end{table*}

\subsection{Main Results}

\textbf{Effectiveness of semi-supervised learning.} As shown in Tab.~\ref{tab:method_comparison}, all semi-supervised methods, using only 25\% labeled data, surpass the supervised baseline trained on 100\% labeled data. This demonstrates the effectiveness of semi-supervised learning with our unlabeled non-swallowing segments.

\textbf{Effectiveness of \sexyname.} As shown in Tab. \ref{tab:method_comparison}, integrating our \sexyname into SemiPose~\cite{xie2021empirical} and G2LCPS \cite{ren2024g2lcps} outperforms existing methods. Compared to SemiPose, \sexyname achieves PCK gains of 2.1\%, 3.9\%, and 2.1\% at 25\%, 50\%, and 100\% supervision, respectively, while notably reducing MED by 20.7\% and 15.9\% at 25\% and 50\% supervision. When combined with G2LCPS, \sexyname further improves PCK by 1.5\%, 2.3\%, and 0.8\%, again accompanied by lower MED, and achieves the best overall performance. These results confirm that our plug-and-play design effectively leverages both labeled and unlabeled data for robust keypoint localization, and can serve as a complementary module to existing methods.

\begin{table}[t]
\caption{Ablation study on the proposed Structure-Aware Learning (SAL) and Structural Representation Consistency (SRC) Learning. PCK@0.02 (\%) and Mean Euclidean Distance (MED, in mm) are reported.}
  \label{tab:ablation_study}
  \vspace{-5pt}
\renewcommand{\arraystretch}{1.0}
\renewcommand{\tabcolsep}{4pt}
\begin{center}
    \begin{tabular}{lcccccc}
    \toprule
    \multirow{2}{*}{Method} & \multicolumn{2}{c}{25\%} & \multicolumn{2}{c}{50\%} & \multicolumn{2}{c}{100\%} \\
    \cmidrule(lr){2-3} \cmidrule(lr){4-5} \cmidrule(lr){6-7}
                            & PCK↑ & MED↓ & PCK↑ & MED↓ & PCK↑ & MED↓ \\
    \midrule
    Baseline~\cite{xie2021empirical}      & 77.2 & 3.63  & 79.0 & 3.14  & 81.9 & 2.65 \\
    Baseline w/ SAL                       & 78.1 & 3.22  & 81.6 & 2.69  & 82.4 & 2.58 \\
    Baseline w/ SRC                        & 78.4 & 3.62  & 81.2 & 2.72  & 82.7 & 2.57 \\
    \sexyname (Ours)                      & \textbf{78.8} & \textbf{2.88} & \textbf{82.1} & \textbf{2.64} & \textbf{83.6} & \textbf{2.48} \\
    \bottomrule
    \end{tabular}
    
\end{center}
\vspace{-15pt}
\end{table}

  
\begin{table}[t]
\caption{Effect of auxiliary models and upsampling strategies on the Structure Parsing Module. PCK@0.02 (\%) and Mean Euclidean Distance (MED, in mm) are reported.}
    \label{tab:SAB}
    \vspace{-10pt}
\renewcommand{\arraystretch}{1.0}
\renewcommand{\tabcolsep}{2.5pt}
\begin{center}
    \begin{tabular}{llcccccc}
    \toprule
    \multirow{2}{*}{Model} & \multirow{2}{*}{Upsample} & \multicolumn{2}{c}{25\%} & \multicolumn{2}{c}{50\%} & \multicolumn{2}{c}{100\%} \\
    \cmidrule(lr){3-4} \cmidrule(lr){5-6} \cmidrule(lr){7-8}
                                     &                             & PCK↑ & MED↓ & PCK↑ & MED↓ & PCK↑ & MED↓ \\
    \midrule
    CLIP    & FeatUp     & 77.8 & 3.73 & 80.5 & 2.86 & 81.9 & 2.74 \\
    DINOv2  & FeatUp     & \textbf{78.1} & 3.22 & \textbf{81.6} & \textbf{2.69} & \textbf{82.4} & \textbf{2.58} \\
    DINOv2  & Bilinear   & 77.8 & \textbf{2.97} & 80.9 & 2.79 & 81.9 & 2.68 \\
    \bottomrule
    \end{tabular}
    
\end{center}
\vspace{-15pt}
\end{table}

  

\begin{table}[t]
\caption{Effect of unshuffle position on the Block Shuffle (BS) operation. ``final" indicates applying unshuffle to the final output, while ``encoder" indicates applying unshuffle to the encoder output.}
    \label{tab:BS}
    \vspace{-10pt}
\renewcommand{\arraystretch}{1.0}
\renewcommand{\tabcolsep}{4pt}
\begin{center}
    \begin{tabular}{lcccccc}
    \toprule
    \multirow{2}{*}{Method} & \multicolumn{2}{c}{25\%} & \multicolumn{2}{c}{50\%} & \multicolumn{2}{c}{100\%} \\
    \cmidrule(lr){2-3} \cmidrule(lr){4-5} \cmidrule(lr){6-7}
                            & PCK↑ & MED↓ & PCK↑ & MED↓ & PCK↑ & MED↓ \\
    \midrule
    w/o BS                              & 77.2 & 3.63 & 79.0 & 3.14 & 81.9 & 2.65 \\
   BS (final)     & 77.7 & \textbf{3.16} & 80.9 & 2.77 & 82.5 & 2.59 \\
   BS (encoder)   & \textbf{78.4} & 3.62 & \textbf{81.2} & \textbf{2.72} & \textbf{82.7} & \textbf{2.57} \\
    \bottomrule
    \end{tabular}
\end{center}
\vspace{-5pt}
\end{table}


\subsection{Ablation Studies}
We evaluate the effect of key components in our \sexyname, using SemiPose~\cite{xie2021empirical} as the semi-supervised baseline.

\textbf{Effect of our proposed Structure-Aware Learning and  Structural Representation Consistency Learning.} Tab.~\ref{tab:ablation_study} shows that both Structure-Aware Learning (SAL) and Structural Representation Consistency  (SRC) Learning contribute to performance gains. SAL enhances keypoint localization by leveraging structural context, while SRCL improves robustness by enforcing consistency in structural representations. Their combination, \sexyname, achieves the best results, with PCK improvements along with notable reductions in MED.






\textbf{Effect of auxiliary models and upsampling strategies of the Structure Parsing Module.} The Structure Parsing Module, part of Structure-Aware Learning, can leverage different pre-trained models as auxiliaries. In addition to DINOv2, we explore CLIP~\cite{radford2021learning}, another large-scale pre-trained model, as an auxiliary model. As shown in Tab.~\ref{tab:SAB}, using CLIP outperforms the baseline but lags behind DINOv2, as it relies on image-text pairs rather than self-supervised learning from image data, limiting its ability to provide adequate structural priors. We also evaluate alternative upsampling techniques, finding that replacing FeatUp with bilinear interpolation reduces performance, highlighting FeatUp's effectiveness in preserving structures while restoring high-resolution spatial details. 



\textbf{Effect of unshuffle position in Structure Representation Consistency Learning.} We investigate the impact of applying the unshuffle operation at different stages: (1) Block Shuffle (BS) with unshuffle on encoder output and (2) BS with unshuffle on final output. Tab.~\ref{tab:BS} shows that both variants improve performance over the baseline (w/o BS). Applying it to the encoder output achieves the best overall results, with consistently higher PCK and lower MED at 50\% and 100\% supervision. Although its MED at 25\% supervision is relatively high, this effect diminishes with more labeled data, and the encoder variant remains clearly superior in most settings. These results indicate that restoring disrupted features at the encoder stage yields stronger spatial and structural representations, making it the preferred choice.






\begin{figure}[t]
    \centering
    \includegraphics[width=0.48\textwidth]{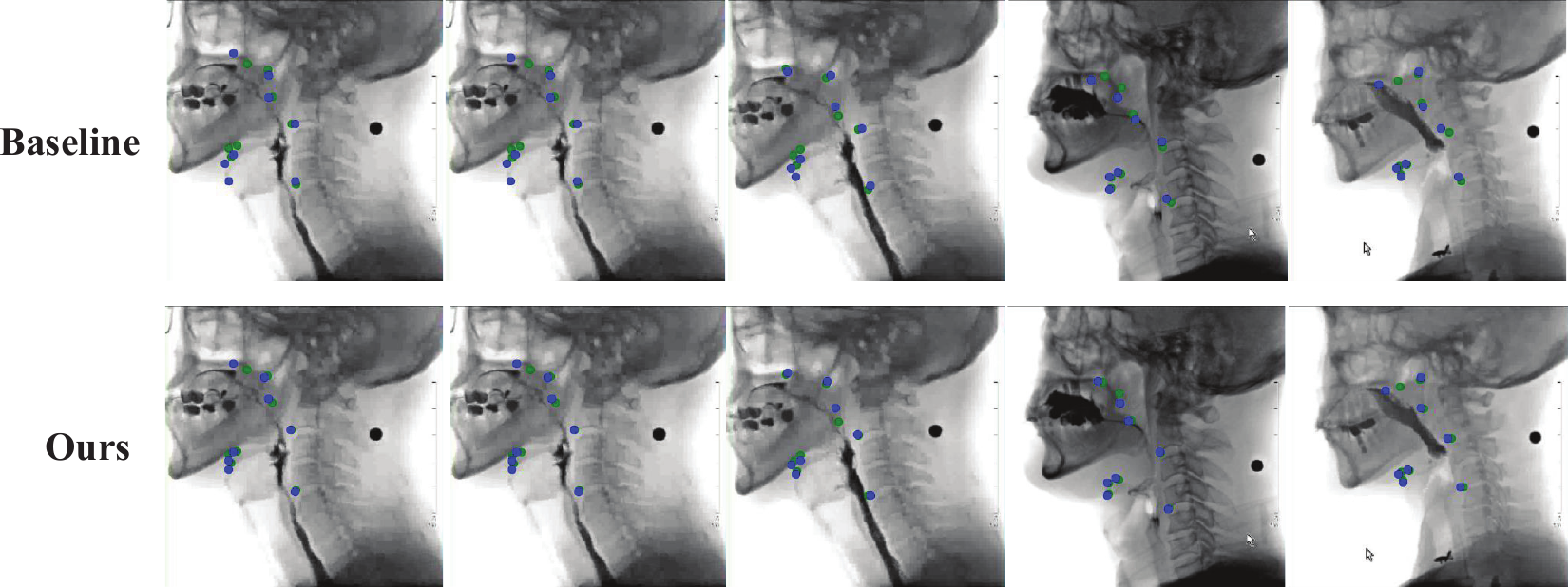}
    \vspace{-5pt}
    \caption{Comparison of keypoint localization results between the baseline method~\protect\cite{xie2021empirical} and our \sexyname under 25\% supervision. Green points represent the ground truth positions, while blue points represent the predicted keypoints.}
    \label{fig:visplus}
    \vspace{-10pt}
\end{figure}

\subsection{Visualization Results}\label{app:Visual}
We visualize the keypoint localization results of the baseline method~\cite{xie2021empirical} and our proposed \sexyname under the 25\% supervision level. As shown in Fig.~\ref{fig:visplus}, our method consistently achieves more accurate keypoint predictions, with the predicted blue points aligning closely with the ground truth green points. From columns 1–3, the baseline fails to perceive the hyoid bone and instead memorizes average training coordinates rather than tracking its true anatomical location. In contrast, \sexyname demonstrates better performance, highlighting the effectiveness of our approach in capturing anatomical structures and improving localization accuracy.

\section{Conclusion}
We introduce the VFSSKep dataset and the \sexyname method for semi-supervised VFSS keypoint localization. By mitigating spatial bias with structure-aware priors and consistency learning, our method learns robust anatomical semantics rather than memorizing coordinates. Experiments show that it outperforms fully supervised methods with only 25\% labeled data, providing a data-efficient solution for clinical kinematic analysis.

\bibliographystyle{IEEEtran}
\bibliography{ref}

\end{document}


\title{Supplementary of \\``Structure-aware Keypoint Localization for \\ Videofluoroscopic Swallowing Study"}


\maketitle


\section{Related Work}\label{app:Related Work}
\subsection{Videofluoroscopic Swallowing Study}
Videofluoroscopic Swallowing Study (VFSS) is the gold standard for diagnosing swallowing disorders \cite{na2019thyroid}, providing dynamic X-ray imaging of the swallowing process. In VFSS, quantitative diagnosis relies on temporal and kinematic analysis. Temporal analysis \cite{lee2019automatic,ruan2023temporal} involves identifying the start and end times of swallowing actions or micro-actions in the video, while kinematic analysis \cite{zhang2021automatic,hsiao2023deep} requires localizing anatomical landmarks to extract kinematic parameters associated with swallowing. Existing studies have explored various approaches for kinematic analysis. For instance, Kellen et al. \cite{kellen2010computer} manually label the hyoid region and use Sobel edge detection to track its motion. Zhang et al. \cite{zhang2021automatic} propose a deep learning-based method to localize keypoints of cervical vertebrae, while Hsiao et al. \cite{hsiao2023deep} develop a deep learning-based approach to localize the hyoid bone and cervical vertebrae. However, these studies focus on limited keypoints (e.g., cervical vertebrae or hyoid regions) and neglect critical areas like the soft palate \cite{logemann2007swallowing}. In contrast, we introduce a new keypoint localization dataset for VFSS, annotating 8 categories of keypoints, including the soft palate, cervical vertebrae, and hyoid regions. Building on this dataset, we propose a semi-supervised keypoint localization framework for VFSS to further improve localization performance.

\subsection{Keypoint Localization}
Keypoint Localization has been widely applied in various domains, including human pose estimation \cite{toshev2014deeppose,carreira2016human,sun2017compositional,tompson2014joint,wei2016convolutional,newell2016stacked,xiao2018simple,sun2019deep,xu2022vitpose}, facial landmark detection \cite{gao2020coarse,mahpod2021facial}. Keypoint localization can be broadly categorized into regression-based methods~\cite{toshev2014deeppose,carreira2016human,sun2017compositional} and heatmap-based methods~\cite{tompson2014joint,wei2016convolutional,newell2016stacked,xiao2018simple,sun2019deep,xu2022vitpose}, depending on the type of supervision signal. Regression-based methods directly predict the 2D coordinates of keypoints by learning the mapping from images to these coordinates. Toshev et al. \cite{toshev2014deeppose} were the first to use deep convolutional networks for pose estimation tasks, employing AlexNet \cite{krizhevsky2012imagenet} as the backbone network and proposing a deep convolutional regressor named DeepPose to predict keypoint coordinates. Carreira et al. \cite{carreira2016human} employed GoogLeNet \cite{szegedy2015going} as the backbone network and introduced an Iterative Error Feedback network, a self-correcting model that progressively improves the prediction of keypoint coordinates by feeding prediction errors back into the input space. However, directly learning the mapping from images to sparse keypoint coordinates is quite challenging, and the performance of such methods often falls short compared to heatmap-based methods. Heatmap-based methods do not directly predict the 2D coordinates of keypoints; instead, they predict heatmaps for each keypoint, where the ground truth heatmap is typically generated by a 2D Gaussian distribution centered at the keypoint location.  Newell et al. \cite{newell2016stacked} proposed the Stacked Hourglass model, whose hourglass-shaped structure enhances the receptive field while minimizing computational demand for pose estimation. Xiao et al. \cite{xiao2018simple} proposed the SimpleBaseline model, which employs ResNet \cite{he2016deep} as the backbone network and adds only a few transposed convolutional layers to generate keypoint heatmaps.
Sun et al. \cite{sun2019deep} recognized the benefits of high-resolution features for keypoint localization and innovatively proposed the High-Resolution Network (HRNet). This model maintains high-resolution feature representations throughout the network, gradually incorporates sub-networks of decreasing resolutions, and integrates their features at multiple scales to form a multi-resolution feature map. In addition to CNN-based methods, transformers have shown strong performance in keypoint localization. For example, ViTPose \cite{xu2022vitpose} leverages self-attention to capture long-range dependencies and achieves superior performance. In this paper, we utilize SimpleBaseline \cite{xiao2018simple} as the main framework of keypoint localization to extract features following \cite{xie2021empirical}. Our method allows for the replacement of the keypoint localization framework with various network architectures (\eg, ViTPose \cite{xu2022vitpose}) in semi-supervised training.

\subsection{Semi-supervised Keypoint Localization}
Semi-supervised keypoint localization employs pseudo-labeling~\cite{radosavovic2018data,wang2022pseudolabeled} and consistency regularization~\cite{xie2021empirical,moskvyak2021semisupervised,Huang_2023_CVPR,ren2024g2lcps} paradigms. DataDistill~\cite{radosavovic2018data} uses ensembled predictions from a teacher model on transformed unlabeled images as pseudo-labels to train a student model, while PLACL~\cite{wang2022pseudolabeled} introduces a reinforcement learning-based strategy for pseudo-label selection. However, pseudo-labeling is often limited by the fixed teacher model pre-trained on labeled data. To address this, Xie \etal~\cite{xie2021empirical} propose consistency regularization with an easy-hard augmentation strategy to avoid the collapsing problem (\ie, predicting all pixels as background). SSKL~\cite{moskvyak2021semisupervised} jointly optimizes keypoint heatmaps and category-level representations using supervised and unsupervised constraints, and Huang \etal~\cite{Huang_2023_CVPR} refine pseudo-labels by ensembling the least inconsistent predictions from two teacher networks. Yu \etal~\cite{yu2024denoising} introduce a denoising scheme to generate reliable pseudo-heatmaps. G2LCPS~\cite{ren2024g2lcps} further enhances semi-supervised keypoint localization with a global filter for selecting unlabeled data and a local filter for removing unreliable pseudo heatmaps. However, these methods ignore anatomical cues and structural relationships vital for keypoint localization in VFSS. Our \sexyname bridges this gap by leveraging structural priors to improve robustness and accuracy.



\section{Dataset Construction Details}
\label{app:dataset_details}

\textbf{Dataset Acquisition Details.} We use two digital fluoroscopic systems to enhance clinical relevance and diversity: \textbf{1)} Lanmage Dynamic Digital Radiography Machine (Athena Plus 7500; Shenzhen Lanmage Medical Technology Co., Ltd., China), accounting for 65\% of the videos, and \textbf{2)} Gastrointestinal X-ray Machine (Toshiba DBA-300; Toshiba Co. Ltd., Tokyo, Japan), accounting for the remaining 35\%. All fluoroscopic videos were recorded at 30 FPS, and frames were converted into grayscale PNG format after extraction.

\textbf{Dataset Statistics.} We collect 288 videos and curate them based on specific criteria, including clear visualization of swallowing phases and diverse swallowing patterns. From these, we extract 13,895 frames from 210 videos, focusing on active swallowing segments. These frames are divided into three subsets: 8,302 images (from 131 videos) for the labeled training set, 1,518 images (from 16 videos) for the validation set, and 4,075 images (from 63 videos) for the test set. We maintain a 6:1:3 ratio for the training, validation, and test sets, ensuring that each subset includes images from distinct individuals to prevent data overlap. The remaining 78 videos constitute the unlabeled dataset. To ensure diversity, both active swallowing and non-swallowing segments are included, with the latter constituting the majority of the duration. Since non-swallowing segments contain relatively static frames, we sample one frame every ten frames, resulting in 39,838 images. After filtering out frames with poor visibility of the oral cavity or throat, as well as those with excessive motion blur, we finalize the unlabeled dataset with 31,536 clean images.

\textbf{Demographic Information.} 
Table~\ref{tab:demographics} summarizes the demographic distribution of subjects in VFSSKep. The dataset covers a wide age range (10–70+ years) and maintains a balanced sex distribution across training, validation, test, and unlabeled subsets, ensuring diversity for model development and evaluation.  

\begin{table}[tb]
\small
\centering
\caption{Demographic distribution of VFSSKep subjects. Values are percentages; $n$: number of subjects.}
\label{tab:demographics}
\renewcommand{\tabcolsep}{8pt}
\begin{tabular}{lcccc}
\toprule
& \begin{tabular}[c]{@{}c@{}}\textbf{Unlabeled}\\($n=62$)\end{tabular}
& \begin{tabular}[c]{@{}c@{}}\textbf{Train}\\($n=28$)\end{tabular}
& \begin{tabular}[c]{@{}c@{}}\textbf{Val}\\($n=9$)\end{tabular}
& \begin{tabular}[c]{@{}c@{}}\textbf{Test}\\($n=28$)\end{tabular} \\
\midrule
Male   & 37.1\% & 39.3\% & 55.6\% & 39.3\% \\
Female & 62.9\% & 60.7\% & 44.4\% & 60.7\% \\
\midrule
Age 10–30 & 9.7\%  & 7.1\%  & 33.3\% & 7.1\%  \\
Age 30–50 & 25.8\% & 25.0\% & 11.1\% & 21.4\% \\
Age 50–70 & 41.9\% & 42.9\% & 44.4\% & 46.4\% \\
Age 70+   & 22.6\% & 25.0\% & 11.1\% & 25.0\% \\
\bottomrule
\end{tabular}
\end{table}

\textbf{Annotation Tool.} We utilized the open-source CVAT tool\footnote{https://github.com/cvat-ai/cvat}. A cross-validation process was implemented where annotations were reviewed by independent experts to ensure consistency.

\textbf{Ethical Approval and Informed Consent.}
All videos were collected from the Third Affiliated Hospital of Sun Yat-Sen University. This study was approved by the hospital's Ethics Committee (approval number: [2018]02-374-01). Informed consent was waived due to the retrospective nature of the study and the use of anonymized data.

\section{More Experimental Results}

\begin{table}[t]
\small
\centering
\caption{An example implementation of the SAB prediction head using ResNet18 as the backbone. The deconvolutional layers (dec$_1$, dec$_2$, dec$_3$) progressively upsample the features, and a final $1\times1$ convolution (pred) produces the output.}
\label{tab:details of SAB}
\renewcommand{\tabcolsep}{8pt}
\begin{tabular}{c c c}
\toprule
Stage   & Architecture                                                                             & Output sizes $C \times H \times W$ \\
\midrule
input   &                                                                                          & $512 \times 8 \times 8$           \\
\cmidrule(lr){1-3}
dec$_1$ & \begin{tabular}[c]{@{}c@{}}deconv $4 \times 4, 256$\\ (stride=2, padding=1)\end{tabular} & $256 \times 16 \times 16$         \\
\cmidrule(lr){1-3}
dec$_2$ & \begin{tabular}[c]{@{}c@{}}deconv $4 \times 4, 256$\\ (stride=2, padding=1)\end{tabular} & $256 \times 32 \times 32$         \\
\cmidrule(lr){1-3}
dec$_3$ & \begin{tabular}[c]{@{}c@{}}deconv $4 \times 4, 256$\\ (stride=2, padding=1)\end{tabular} & $256 \times 64 \times 64$         \\
\cmidrule(lr){1-3}
pred    & \begin{tabular}[c]{@{}c@{}}conv $1 \times 1, 384$\\ (stride=1, padding=0)\end{tabular}   & $384 \times 64 \times 64$         \\
\bottomrule
\end{tabular}
\end{table}

\subsection{More Implementation Details}
Following SemiPose~\cite{xie2021empirical}, we employ SimpleBaseline~\cite{xiao2018simple} as the keypoint localization network, utilizing ResNet18~\cite{he2016deep} as its backbone (shared encoder).
We train our model on 8 TITAN Xp GPUs using PyTorch. To compute MED across devices, we normalize each image using its original pixel spacing and resize it to $256 \times 256$, resulting in a unified scale of 0.6 mm/pixel. For the prediction head of Structure-Aware Branch (SAB), we employ a straightforward yet efficient design. It consists of three deconvolutional layers that enhance the resolution of the feature maps, along with a convolutional layer that generates the feature maps. The structure of the deconvolutional layers is inspired by the decoder architecture of SimpleBaseline. The detailed architecture is presented in Table \ref{tab:details of SAB}, with the input coming from the output of the final convolutional stage in ResNet.
Our Structure Parsing Module (SPM) extracts low-resolution features via a fixed DINOv2 and passes them, along with the original image, through a 4-layer Joint Bilateral Upsampler (JBU) from FeatUp~\cite{fu2024featup}. Each JBU layer upsamples by a factor of 2, achieving 16× upsampling for high-resolution output.
In our experiments,we train for 100 epochs using the Adam optimizer with an initial learning rate of $10^{-3}$, decayed to $10^{-4}$ at epoch 60 and $10^{-5}$ at epoch 90. The batch size is 64. For data augmentation, we apply easy augmentation (e.g., Affine Transformation with rotation angles sampled from $[-30^\circ, 30^\circ]$) to the teacher network and hard augmentation (e.g., Affine Transformation with rotation angles sampled from $[-60^\circ, 60^\circ]$ combined with our proposed Block Shuffle) to the student network. Following SemiPose~\cite{xie2021empirical}, we set the supervised loss weight $\lambda_1$ and unsupervised loss weight $\lambda_2$ to 0.5 and 1.0, respectively. In addition to the default hyperparameters, we set the feature alignment loss weight $\lambda_3 = 10^{-3}$. The similarity threshold $\alpha$ is 0.85, and the block size of Block Shuffle is $64 \times 64$. We analyze the effects of these hyperparameters in hyperparameter analyses.

\begin{table}[t]
\small
\caption{Effect of loss weight $\lambda_3$ for feature alignment loss. PCK@0.02 (\%) and Mean Euclidean Distance (MED, in mm) are reported.}
\label{tab:loss_weight_hyperparameters}
\renewcommand{\arraystretch}{1.0}
\renewcommand{\tabcolsep}{6pt}
\begin{center}
\begin{tabular}{lcccccc}
\toprule
\multirow{2}{*}{$\lambda_3$} & \multicolumn{2}{c}{25\%} & \multicolumn{2}{c}{50\%} & \multicolumn{2}{c}{100\%} \\
\cmidrule(lr){2-3} \cmidrule(lr){4-5} \cmidrule(lr){6-7}
                             & PCK↑ & MED↓ & PCK↑ & MED↓ & PCK↑ & MED↓ \\
\midrule
10$^{-1}$                   & 69.2 & 3.60 & 73.7 & 4.08 & 75.3 & 3.14 \\
10$^{-2}$                   & 78.1 & 3.56 & 80.0 & 2.90 & 81.1 & 2.75 \\
10$^{-3}$                   & \textbf{78.8} & \textbf{2.88} & \textbf{82.1} & \textbf{2.64} & \textbf{83.6} & \textbf{2.48} \\
10$^{-4}$                   & 78.4 & 3.62 & 80.6 & 3.32 & 82.4 & 2.56 \\
\bottomrule
\end{tabular}

\end{center}
\end{table}

\begin{table}[t]
\small
\renewcommand{\arraystretch}{1.0}
\renewcommand{\tabcolsep}{6.5pt}
\begin{center}
\caption{Effect of the similarity threshold $\alpha$ in the cosine similarity loss $\mathcal{L}_{cos}$. PCK@0.02 (\%) and Mean Euclidean Distance (MED, in mm) are reported.}
\label{tab:alpha_thresholds}
\begin{tabular}{lcccccc}
\toprule
\multirow{2}{*}{$\alpha$} & \multicolumn{2}{c}{25\%} & \multicolumn{2}{c}{50\%} & \multicolumn{2}{c}{100\%} \\
\cmidrule(lr){2-3} \cmidrule(lr){4-5} \cmidrule(lr){6-7}
                         & PCK↑ & MED↓ & PCK↑ & MED↓ & PCK↑ & MED↓ \\
\midrule
0.7                      & 78.4 & 2.94 & 82.0 & 2.87 & 83.4 & 2.49 \\
0.85                     & \textbf{78.8} & \textbf{2.87} & \textbf{82.1} & \textbf{2.64} & \textbf{83.6} & \textbf{2.48} \\
1.0                      & 78.7 & 3.66 & 81.6 & 2.87 & 82.8 & 2.58 \\
\bottomrule
\end{tabular}

\end{center}
\vspace{-10pt}
\end{table}

\begin{table}[t]
\small
\renewcommand{\arraystretch}{1.0}
\renewcommand{\tabcolsep}{6pt}
\begin{center}
\caption{Effect of input block size on the Block Shuffle operation. PCK@0.02 (\%) and Mean Euclidean Distance (MED, in mm) are reported.}
\label{tab:shuffle_block_size}
\begin{tabular}{lcccccc}
\toprule
\multirow{2}{*}{Size} & \multicolumn{2}{c}{25\%} & \multicolumn{2}{c}{50\%} & \multicolumn{2}{c}{100\%} \\
\cmidrule(lr){2-3} \cmidrule(lr){4-5} \cmidrule(lr){6-7}
                      & PCK↑ & MED↓ & PCK↑ & MED↓ & PCK↑ & MED↓ \\
\midrule
32                   & 78.5 & 2.95 & 81.8 & 2.75 & 82.6 & 2.68 \\
64                   & \textbf{78.8} & \textbf{2.87} & \textbf{82.1} & \textbf{2.64} & \textbf{83.6} & \textbf{2.48} \\
128                  & 78.4 & 3.65 & 81.4 & 3.13 & 82.3 & 2.60 \\
\bottomrule
\end{tabular}
\end{center}
\vspace{-10pt}
\end{table}

\subsection{Hyperparameter Analyses}

\textbf{Effect of loss weight $\lambda_3$.} Tab.~\ref{tab:loss_weight_hyperparameters} shows the model's performance with varying loss weights $\lambda_3$ for the auxiliary task. A high weight (\eg, $\lambda_3 = 10^{-1}$) overemphasizes the auxiliary task, degrading performance on the primary task. A low weight (\eg, $\lambda_3 = 10^{-4}$) diminishes the auxiliary task's impact. The best performance occurs at $\lambda_3 = 10^{-3}$, balancing both tasks.


\textbf{Effect of similarity threshold $\alpha$.} 
Tab.~\ref{tab:alpha_thresholds} shows how different similarity threshold values $\alpha$ affect the calculation of the Structure-Aware Learning loss. When we set $\alpha$ a low threshold (\eg, 0.7), the model excludes a large proportion of features from the Structure-Aware Learning loss, potentially hindering its alignment learning. Conversely, when we set a high threshold (\eg, 1.0), the model includes all features, even those already well-optimized, which may lead to overfitting. The optimal performance is achieved at $\alpha = 0.85$, balancing relevant feature inclusion for better generalization.   

\textbf{Effect of input block size on the Block Shuffle operation.} The Block Shuffle operation rearranges image blocks before inputting them into the encoder. A small block size (32$\times$32) causes excessive shuffling, disrupting local spatial relationships and lowering performance. A large block size (128$\times$128) provides insufficient shuffling, failing to prevent overfitting. As shown in Tab.~\ref{tab:shuffle_block_size}, the optimal block size of 64$\times$64 strikes a balance, preserving spatial structures while enabling effective shuffling, resulting in the best performance across all supervision levels.


\begin{table}[t]
\small
\caption{Effect of different backbones.}
\label{tab:backbone}
\renewcommand{\arraystretch}{1.0}
\renewcommand{\tabcolsep}{1.8pt}
\begin{center}
    \begin{tabular}{lcccccccc}
    \toprule
    \multirow{2}{*}{Backbone} & \multirow{2}{*}{Param} & \multirow{2}{*}{FLOPs} & \multicolumn{2}{c}{25\%} & \multicolumn{2}{c}{50\%} & \multicolumn{2}{c}{100\%} \\
    \cmidrule(lr){4-5} \cmidrule(lr){6-7} \cmidrule(lr){8-9}
                             &  &  & PCK↑ & MED↓ & PCK↑ & MED↓ & PCK↑ & MED↓ \\
    \midrule
    ResNet18  & 15.6M & 2.4G  & 78.8 & 2.88 & 82.1 & 2.64 & \textbf{83.6} & \textbf{2.48} \\
    ViT-B  & 87.6M & 21.6G & \textbf{79.9} & \textbf{2.74} & \textbf{82.3} & \textbf{2.55} & 83.5 & 2.51 \\
    \bottomrule
    \end{tabular}
\end{center}
\end{table}

\subsection{Extension Experiments}
\textbf{Effect of different backbones.} As shown in Tab.~\ref{tab:backbone}, employing a stronger backbone ViT-B \cite{dosovitskiy2020image} (adopted from ViTPose \cite{xu2022vitpose}) yields better performance under limited supervision (25\% and 50\%), demonstrating its advantage in capturing global structural information. Interestingly, when full supervision is available, the lightweight ResNet18 slightly outperforms ViT-B. We attribute this to the relatively modest dataset size, where the strong capacity of ViT-B is not fully exploited and may even lead to overfitting, while ResNet18 benefits from the inherent inductive bias of convolutional networks that better capture local structural patterns. 
Considering the significantly higher computational and memory cost of ViT-B (87.6M/21.6G vs. 15.6M/2.4G), we mainly adopt ResNet18, as it provides a better balance between accuracy and efficiency for resource-constrained medical scenarios.

\section{More Discussions}
\subsection{Clinical Implications}

Accurate and efficient keypoint localization in VFSS has direct clinical value. Anatomical landmarks underpin quantitative kinematic measures, such as hyoid displacement and soft palate elevation, which are essential for dysphagia assessment. By reducing reliance on labor-intensive annotations, \sexyname lowers deployment barriers in resource-constrained clinical settings. Its plug-and-play design further ensures compatibility with existing pipelines, facilitating seamless integration into current workflows. Moreover, by leveraging both swallowing and non-swallowing phases, \sexyname shows potential for automated pre-screening, helping reduce clinician workload and improve efficiency.

\begin{figure*}[t]
    \centering
    \includegraphics[width=0.98\linewidth]{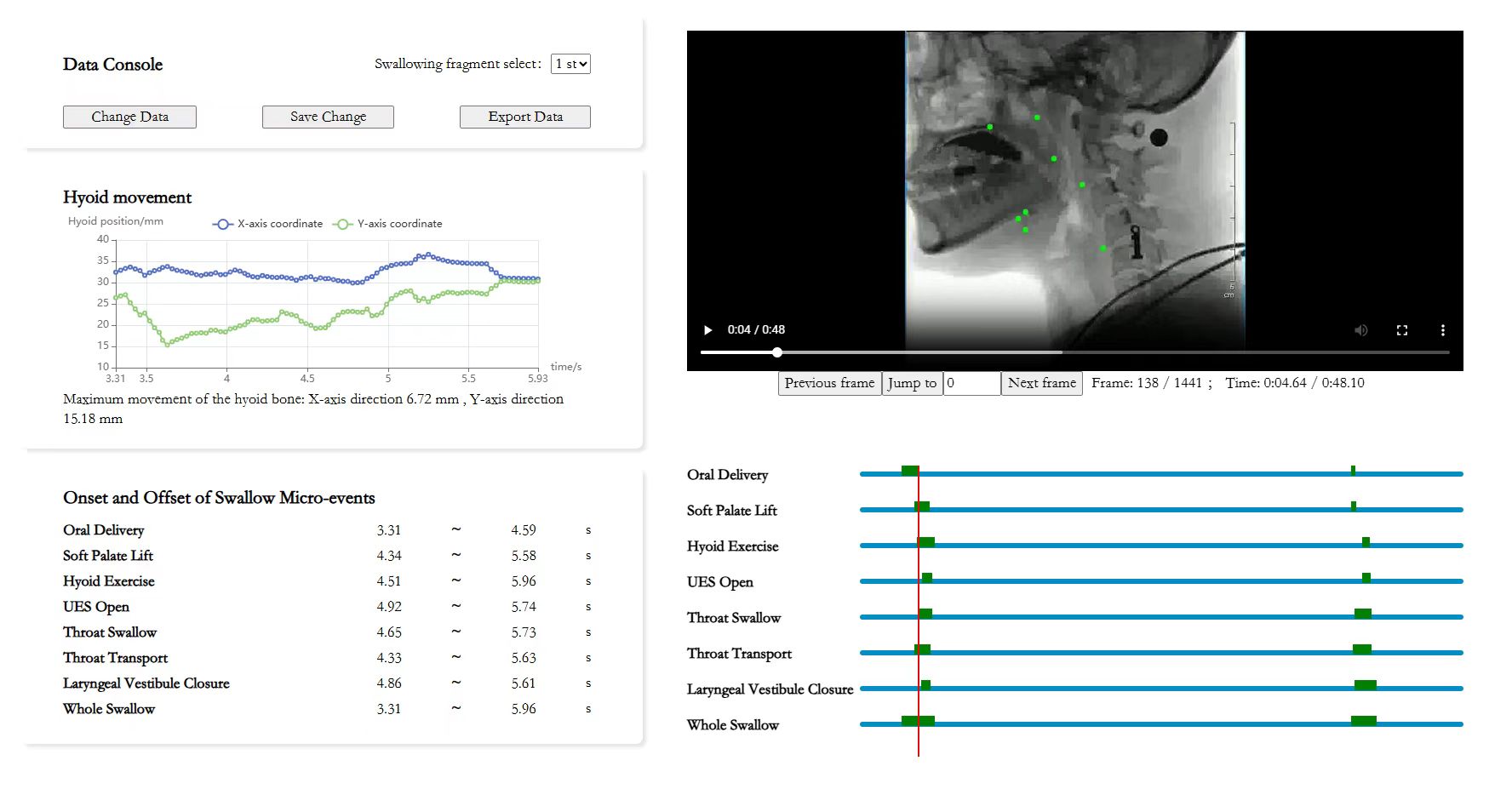}
    \vspace{-5pt} 
    \caption{Partial screenshot of the clinical prototype system.}
    \label{fig:rebuttal}
\end{figure*}

To bridge the gap to real-world application, we have developed a clinician-in-the-loop prototype system integrating our framework (see Fig.~\ref{fig:rebuttal}). The system automatically extracts kinematic parameters (e.g., hyoid trajectories) and temporal metrics (e.g., swallow micro-events) to support diagnostic workflows. While large-scale deployment requires formal ethical approval, this prototype demonstrates the practical clinical relevance of our method.

\subsection{Limitations and Future Directions}
Despite the above contributions, there remain opportunities for further improvement. First, while VFSSKep provides more comprehensive annotations than existing datasets, its scale is still modest compared to large benchmarks in computer vision. Expanding the dataset in size and diversity, including multi-center data collection, would enhance generalizability. Second, our framework currently operates on static frames without explicitly modeling temporal swallowing dynamics, which are known to yield additional diagnostic insights. Incorporating temporal modules or sequence-based architectures represents a promising direction to further strengthen clinical applicability.

\bibliographystyle{IEEEtran}
\bibliography{ref}